\documentclass[10pt, a4paper]{article}

\usepackage{lrec2026} 
\usepackage{multirow}
\usepackage{booktabs}
\usepackage{graphicx}
\newcommand{\todo}[1]{}
\usepackage{amsmath}
\usepackage{subcaption}
\usepackage{url}
\usepackage{hyperref}
\usepackage{textcomp}
\usepackage[table]{xcolor}
\usepackage{bbding}
\renewcommand{\todo}[1]{{\color{red} [TODO: {#1}]}}

\usepackage{titlesec}
\titlespacing*{\subsection}{0pt}{1.6ex plus 0.3ex minus 0.2ex}{0.9ex plus 0.2ex}
\titlespacing*{\section}{0pt}{2.2ex plus 0.4ex minus 0.3ex}{1.3ex plus 0.3ex}

\title{When Consistency Becomes Bias: Interviewer Effects in Semi-Structured Clinical Interviews}

\name{%
  \begin{tabular}[t]{c}
    Hasindri Watawana$^{1,2}$, Sergio Burdisso$^{1}$, Diego A. Moreno-Galván$^{3}$\\
    Fernando Sánchez-Vega$^{3}$, A. Pastor López-Monroy$^{3}$, Petr Motlicek$^{1,4}$,\\
    Esaú Villatoro-Tello$^{1}$
  \end{tabular}
}

\address{
  $^1$ Idiap Research Institute, Switzerland\\ 
  {\tt \{hwatawana,sburdisso,pmotlicek,evillatoro\}@idiap.ch} \\[0.4em]
  $^2$ EPFL, Switzerland\\[0.4em]
  $^3$ Centro de Investigación en Matemáticas (CIMAT), Mexico\\
  {\tt \{diego.moreno,fernando.sanchez,pastor.lopez\}@cimat.mx} \\[0.4em]
  $^4$ Brno University of Technology, Czech Republic\\
}

\abstract{
Automatic depression detection from doctor–patient conversations has gained momentum thanks to the availability of public corpora and advances in language modeling. However, interpretability remains limited: strong performance is often reported without revealing what drives predictions. We analyze three datasets—ANDROIDS, DAIC-WOZ, and E-DAIC—and identify a systematic bias from interviewer prompts in semi-structured interviews. Models trained on interviewer turns exploit fixed prompts and positions to distinguish depressed from control subjects, often achieving high classification scores without using participant language. Restricting models to participant utterances distributes decision evidence more broadly and reflects genuine linguistic cues. While semi-structured protocols ensure consistency, including interviewer prompts inflates performance by leveraging script artifacts. Our results highlight a cross-dataset, architecture-agnostic bias and emphasize the need for analyses that localize decision evidence by time and speaker to ensure models learn from participants’ language. Our code is available at: \url{https://github.com/idiap/bias_in_daic-woz/tree/main/LREC_2026}.
\\ \newline  }

\begin{document}

\maketitleabstract

\section{Introduction}
\label{secc:Introduction}

Language has long been established as a powerful indicator of personality, socio-emotional state, and mental health~\cite{pennebaker2003psychological, tackman2019depression}. This insight has spurred a rich body of work at the intersection of AI, natural language processing, and clinical psychology, showing that structured interviews and written responses can reveal important aspects of cognitive and behavioral functioning, particularly for automatic depression detection~\cite{malandrakis2015therapy, villatoroEtAl, villatorotello21_interspeech}.

While a growing body of recent work trains automatic depression detection models on both participant responses and interviewer prompts~\cite{chen-naacl-2024,agarwal2024,milintsevich2023towards,shen2022automatic}, it remains unclear how much each contributes to model performance. This study asks: \emph{to what extent can a model classify a participant as depressed using only the interviewer’s questions?} Surprisingly, across three datasets (ANDROIDS, DAIC-WOZ, E-DAIC) and two model families, \emph{interviewer-only} (\textit{I}) models often match or outperform \emph{participant-only} (\textit{P}) models. This does not imply that interviewer questions are clinically more informative; instead, qualitative analyses show that \textit{I}-models exploit systematic shortcuts tied to the interview script, focusing on recurring prompts and question positions—a phenomenon we term \emph{prompt-induced bias}.  

Our main contributions are threefold: (1) We quantify prompt-induced bias across multiple datasets, showing that \emph{I}-models can outperform \emph{P}-models. (2) We show that this effect is model-agnostic by reproducing it with both graph convolutional networks and transformers. (3) We provide qualitative analyses that reveal how \emph{I}-models concentrate on narrow interview segments while \emph{P}-models distribute evidence more broadly. These findings highlight the importance of carefully handling interviewer prompts and verifying that models truly leverage participant language rather than spurious cues.



\section{Datasets}
\label{secc:Dataset}

Within clinical practice, the initial assessment of mental illness is a semi-structured interview in which clinicians pose a standardized yet open-ended sequence of questions to elicit symptoms, history, and functioning. This protocol balances consistency with flexibility, enabling reliable assessment while allowing patients to elaborate in their own words. The depression corpora we study are designed to simulate such standardized screening protocols for identifying people at risk: an interviewer (human or virtual) delivers a controlled set of prompts to ensure replicability and coverage of diagnostic cues, while participants respond freely, producing language that can be analyzed for clinical indicators.

\subsection{DAIC-WOZ}
\label{subsecc:daicwoz}

The Distress Analysis Interview Corpus – Wizard of Oz (DAIC-WOZ) dataset ~\citelanguageresource{gratch2014distress} contains semi-structured clinical interviews in North American English conducted by “Ellie”, an animated virtual interviewer controlled by a human in another room (“Wizard-of-Oz” setup where a user interacts with a mock interface controlled, to some degree, by a person). DAIC-WOZ is multimodal—audio, video, and manual transcripts—and includes PHQ-8 depression ratings for each participant~\cite{kroenke2009phq}. Ellie’s design emphasizes replicability and consistency: she draws from a finite repertoire of 191 prompts spanning general questions (e.g., lifestyle and sleep), neutral backchannels, positive/negative empathic responses, surprise tokens, continuation prompts (e.g., “could you tell me more?”), and miscellaneous control items. This controlled prompting is intended to elicit behaviors associated with depression and related symptoms while reducing variability attributable to the interviewer. DAIC-WOZ comprises 189 subjects split into 107 train, 35 development, and 47 test interviews. 

\subsection{E-DAIC}
\label{subsecc:edaic}

E-DAIC~\citelanguageresource{devault2014simsensei} is an extension of DAIC-WOZ that preserves the same interview format while scaling the collection. The key difference is that the virtual interviewer used in E-DAIC is fully automatic in contrast to the human-controlled Ellie in DAIC-WOZ. While DAIC-WOZ distributes complete two-speaker transcripts (interviewer and participant), E-DAIC provides only the participant side. Therefore, for our experiments, we prepare automatic transcripts for E-DAIC using WhisperX ~\cite{bain2023whisperxtimeaccuratespeechtranscription} pipeline. The process is explained in Section ~\ref{secc:method}. We corrected several mislabeled subjects after verifying inconsistencies between PHQ scores and the provided binary labels, an issue noted previously by others~\cite{ali2025leveragingaudiotextmodalities}. The official split sizes are 163 train, 56 development, and 56 test interviews.

\subsection{ANDROIDS}
\label{subsecc:androids}

ANDROIDS \citelanguageresource{tao23_interspeech} is an Italian speech corpus for depression detection collected “in the wild” using laptop microphones. Each participant was recorded in two tasks: a Reading Task (RT)—everyone reads the same short, simple story to reduce literacy/education effects—and an Interview Task (IT) with spontaneous speech, where the interviewer is instructed to ask only minimal prompts, yielding semi-structured but low-intervention interactions. In this work we focus only on the data from the IT task. It comprises 116 native-Italian participants (64 depressed, 52 controls), with controls matched to the depressed cohort by gender, age, and education to minimize demographic confounds. Interviews are manually segmented into turns; diagnostic labels (“depressed” vs. “control”) come from clinicians following DSM-5 criteria.\footnote{Diagnostic and Statistical Manual of Mental Disorders, Fifth Edition \cite{apa2013dsm5}.} The release includes audio and acoustic features, but no ground-truth transcripts.  Therefore, we generated WhisperX transcripts for ANDROIDS (see Section \ref{secc:method}). 

\section{Methodology}
\label{secc:method}
\begin{table*}[t]
    \centering
    \small
    \setlength{\tabcolsep}{6pt}
    \begin{tabular}{
        l        
        c c      
        c c c    
        c c c    
        c c c    
    }
    \toprule
    \multirow{2}{*}{\textbf{Model}} &
    \multicolumn{2}{c}{\textbf{Source}} &
    \multicolumn{3}{c}{\textbf{ANDROIDS}} &
    \multicolumn{3}{c}{\textbf{DAIC-WOZ}} &
    \multicolumn{3}{c}{\textbf{E-DAIC$^\dag$}} \\
    \cmidrule(lr){2-3} \cmidrule(lr){4-6} \cmidrule(lr){7-9} \cmidrule(lr){10-12}
    & \textit{P} & \textit{I} &
      \textit{Avg} & \textit{D} & \textit{C} &
      \textit{Avg} & \textit{D} & \textit{C} &
      \textit{Avg} & \textit{D} & \textit{C} \\
    \midrule

    \cite{burdisso23_interspeech} & \Checkmark & &
      -- & -- & -- &
      0.84 & 0.80 & 0.89 &
      0.80 & 0.67 & 0.94 \\

    \cite{ilias24_interspeech} & \Checkmark & \Checkmark &
      0.93 & -- & -- &
      -- & -- & -- &
      -- & -- & -- \\

    \cite{Borraccino2025} & \Checkmark & \Checkmark &
      0.92 & 0.93 & 0.9 &
      -- & -- & -- &
      -- & -- & -- \\

    \cite{milintsevich2023towards} & \Checkmark & \Checkmark &
      -- & -- & -- &
      0.81 & -- & -- &
      -- & -- & -- \\
    \midrule

    \textit{P}-Longformer & \Checkmark & &
      0.79 & 0.82 & 0.76 &
      0.71 & 0.61 & 0.81 &
      \textbf{0.67} & \textbf{0.56} & 0.79 \\

    \textit{\textbf{I}}-Longformer & & \Checkmark &
     \underline{{\textbf{0.98}}} & \underline{\textbf{0.98}} & \underline{\textbf{0.98}} &
      \textbf{0.73} & \textbf{0.64} & \textbf{0.83} &
      0.65 & 0.50 & \textbf{0.80} \\
    \cmidrule(r){4-12}

    \textit{P}-GCN & \Checkmark & &
      0.93 & 0.95 & 0.92 &
      0.85 & 0.81 & 0.88 &
      0.70 & 0.54 & 0.86 \\

    \textit{\textbf{I}}-GCN & & \Checkmark &
      \textbf{0.97} & \textbf{0.97} & \underline{\textbf{0.98}} &
      \underline{\textbf{0.88}} & \underline{\textbf{0.85}} & \underline{\textbf{0.91}} &
      \underline{\textbf{0.74}} & \underline{\textbf{0.57}} & \underline{\textbf{0.90}} \\

    \bottomrule
    \end{tabular}
\caption{Development-set F$_1$ scores on ANDROIDS, DAIC-WOZ, and E-DAIC. For each dataset, we report macro-average (\textit{Avg}), Depressed (\textit{D}), and Control (\textit{C}) F$_1$. ANDROIDS scores are reported as 5-fold averages; DAIC-WOZ and E-DAIC use the official development split. Sources: participant (\textit{P}) and interviewer (\textit{I}) text. Upper block: text-only baselines from prior work. \textbf{Bold} = best in group; \underline{\textbf{underlined}} = best overall text-only result. $\boldsymbol\dag$: E-DAIC prior-work results use gold participant transcripts and are not directly comparable.}
    \label{tab:main_results_three_datasets_dev}
\end{table*}

\noindent \textbf{Data Preparation}
Our study is based on the text modality. DAIC-WOZ provides complete transcripts including both interviewer and participant utterances. ANDROIDS doesn't provide ground-truth transcripts, and E-DAIC releases transcripts for the participant side only. For ANDROIDS and E-DAIC, we therefore built complete textual data as follows: using participant utterance timestamps provided in the metadata, we derived complementary (non-overlapping) interviewer timestamps, extracted interviewer and participant audio clips separately using the timestamp data, and generated automatic transcripts with the WhisperX ASR pipeline~\cite{bain2023whisperxtimeaccuratespeechtranscription} (Whisper large-v3~\cite{radford2022robustspeechrecognitionlargescale} with a faster-whisper backend).\footnote{This pipeline yields a WER of 15.3\% on the E-DAIC participant side, evaluated against the originally provided transcripts as ground truth.} To have a fair comparison between the impact of using interviewer vs. participant utterances on automatic depression detection, we do not use the E-DAIC participant gold transcripts and instead rely on matched ASR transcripts for both speakers. \\

\noindent \textbf{Models}  
We evaluate two architectures to test if prompt effects are model-agnostic: (i) Longformer~\cite{beltagy2020longformer}, a transformer with sparse attention suitable for long documents such as interviews, and (ii) GCN~\cite{burdisso23_interspeech}, a graph-based model with word and document nodes.
Using these two types of models allows us to analyze results from both a contextualized, semantic transformer and a keyword-focused GCN. \\
\noindent
\textbf{$\bullet$ Longformer:} A linear classification head is added on top of the pre-trained Longformer-BERT to classify using the \emph{[CLS]} token; both the encoder and head are fine-tuned on the training split. Experiments are conducted with both longformer-mini-1024 and longformer-base-4096 but only the best-performing configuration is reported.

\noindent
\textbf{$\bullet$ GCN:}
We use the two-layer $\omega$-GCN of \citet{burdisso23_interspeech}, which represents each corpus as a graph with word and document (interview) nodes. Representations evolve through three stages: (i) an initial one-hot layer, (ii) a latent embedding after the first convolution, and (iii) a two-dimensional output after the second convolution, corresponding to depressed and control probabilities. Because words and documents share the same embedding space, the final layer provides class probabilities for both interviews and individual words, offering an interpretability handle that identifies which words—and interviewer prompts—serve as discriminative evidence (see Section~\ref{secc:results}).

\

\noindent \textbf{Quantifying Interviewer Bias}
Our approach builds on \citet{burdisso2024daic}, who analyzed prompt-induced bias within the DAIC-WOZ corpus. We extend this line of work in three ways. First, we generalize the analysis across multiple clinical interview corpora—ANDROID and E-DAIC—to test whether the interviewer-related bias persists beyond a single dataset. Second, while \citet{burdisso2024daic} focused on development splits, we also quantify the impact of interviewer bias on held-out test data, providing a clearer estimate of its effect on reported model performance. Third, we evaluate the bias using automatically generated transcriptions, as manual interviewer transcripts are unavailable for ANDROID and only partially available for E-DAIC. 

This setup allows us to assess whether the bias remains detectable under more realistic, automatically transcribed conditions. For each architecture, we train and evaluate two model variants:
\begin{itemize}
    \item \emph{participant-only} (\textit{P}): trained and evaluated using only the participant’s responses
    \item \emph{interviewer-only} (\textit{I}): trained and evaluated using only the interviewer’s prompts
\end{itemize}


These two model variants allow us to directly investigate our main research question by comparing performance when models have access only to participant responses (\textit{P}) versus only to interviewer prompts (\textit{I}).
This comparison quantifies the extent to which interviewer questions carry implicit diagnostic information, revealing potential bias in automatic depression detection models.

\section{Experiments and Results}
\label{secc:Experiments}

\begin{figure*}[t]
  \centering
  \begin{subfigure}[t]{0.5\textwidth}
    \centering
    \includegraphics[width=\linewidth]{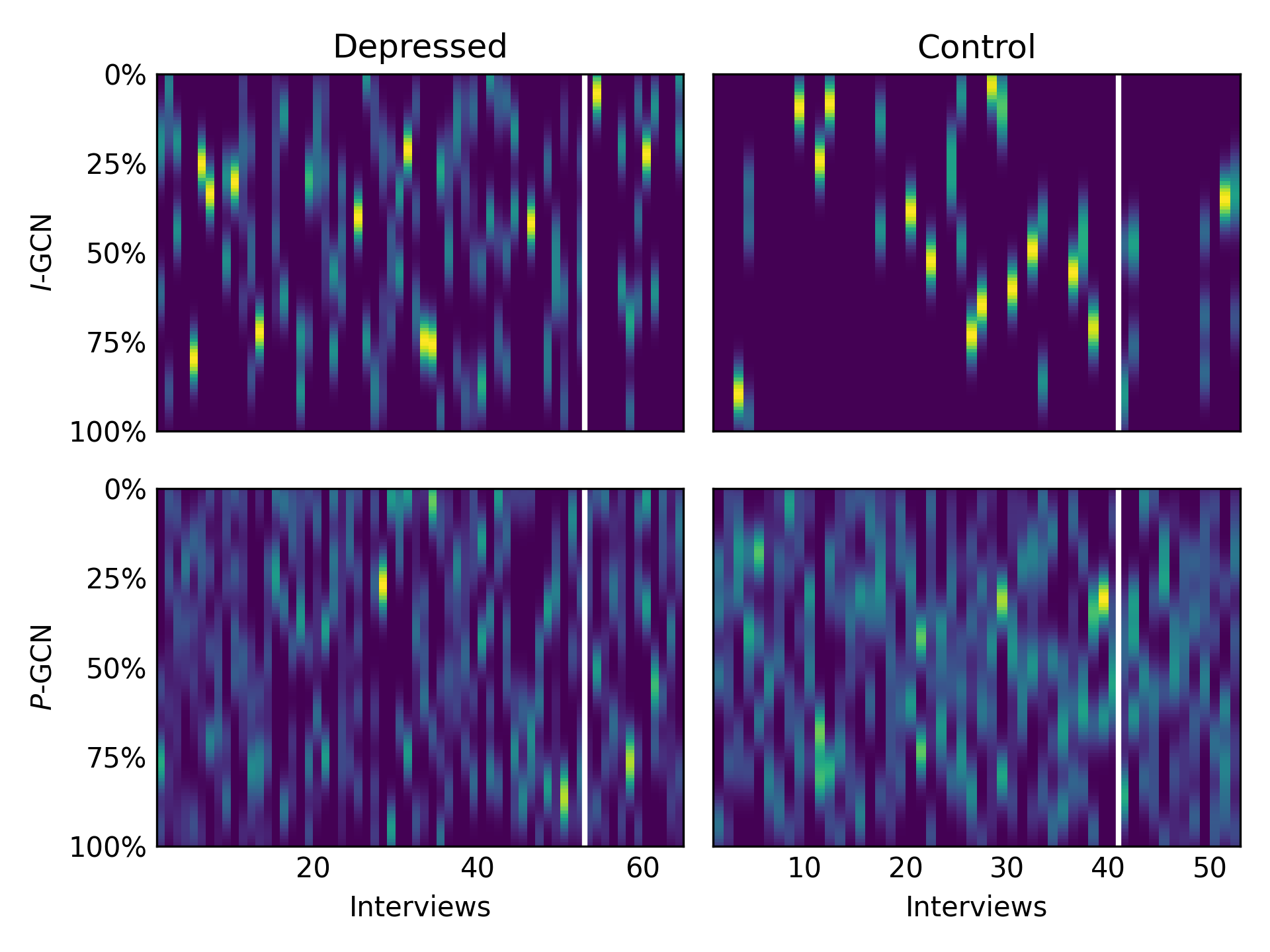}
    \caption{ANDROIDS}
    \label{fig:androids-heatmap}
  \end{subfigure}\hfill
  \begin{subfigure}[t]{0.5\textwidth}
    \centering
    \includegraphics[width=\linewidth]{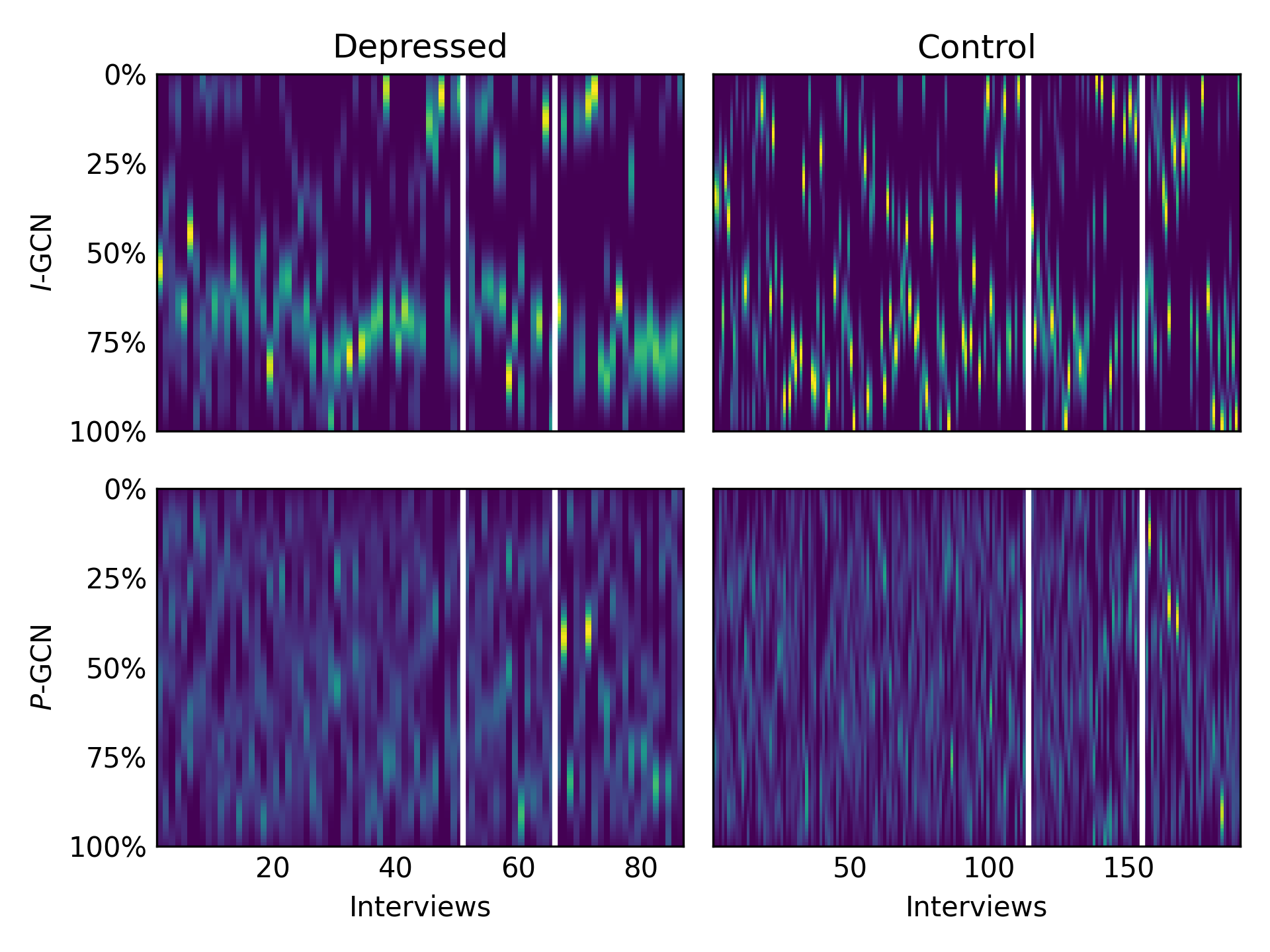}
    \caption{E-DAIC}
    \label{fig:edaic-heatmap}
  \end{subfigure}
  \caption{Temporal heatmaps comparing keyword evidence learned by interviewer-only (\textit{I}, top) vs. participant-only (\textit{P}, bottom) models across interviews in the ANDROIDS and E-DAIC datasets. Each column represents one interview. The y-axis corresponds to the normalized interview timeline, where 0\% marks the beginning of the interview and 100\% marks its end. White vertical lines denote split boundaries (train/dev/test for E-DAIC; train/dev only for ANDROIDS). The ANDROIDS plot is shown for Fold~1.}
  \label{fig:heatmaps-side-by-side}
\end{figure*}

Main results are reported in Table \ref{tab:main_results_three_datasets_dev}. For ANDROIDS five-fold averages are reported; for DAIC-WOZ and E-DAIC the official dev split is used. Across all three corpora we train two variants per architecture—participant-only (\textit{P}-GCN and  \textit{P}-Longformer) and interviewer-only (\textit{I}-GCN and \textit{I}-Longformer). For each GCN model variant, we optimize and evaluate results using multiple setups (e.g. using all vocabulary, only top vocabulary etc), and record the best performance on dev split. Similarly for Longformer, best dev performance among longformer-mini-1024 and longformer-base-4096 is reported.

On DAIC-WOZ, the interviewer side is consistently stronger than the participant side for both model families (P-Longformer 0.71 → I-Longformer 0.73; P-GCN 0.85 → I-GCN 0.88). This reproduces prior findings reported in \cite{burdisso2024daic}, that Ellie’s prompts provide highly discriminative shortcuts, and shows the effect is not tied to a specific architecture.

On ANDROIDS, the bias is even more pronounced: I-Longformer achieves 0.98 macro-F\textsubscript{1} on dev gaining a 19\% advantage over P-Longformer. I-GCN outperforms P-GCN by 4\%. Despite being a different language and collection setting, interviewer prompts again dominate, reinforcing that the advantage arises from interview structure. This bias is evident on E-DAIC with the GCN models; with Longformer, the \textit{P} and \textit{I} variants perform comparably.

\begin{table}[t]
    \centering
    \small
    \setlength{\tabcolsep}{6pt}
    \begin{tabular}{
        l@{~}      
        c@{~~}c@{~~}c  
        c@{~~}c@{~~}c  
    }
    \toprule
    \multirow{2}{*}{\textbf{Model}} &
    \multicolumn{3}{c}{\textbf{DAIC\textendash WOZ}} &
    \multicolumn{3}{c}{\textbf{E\textendash DAIC}} \\
    \cmidrule(lr){2-4}\cmidrule(lr){5-7}
    & \textit{Avg} & \textit{D} & \textit{C} &
      \textit{Avg} & \textit{D} & \textit{C} \\
    \midrule
    \textit{P}-Longformer          & \textbf{0.68} & \textbf{0.54} & \textbf{0.82} & 0.40 & 0.17 & 0.62 \\
    \textit{\textbf{I}}-Longformer & 0.53 & 0.27 & 0.78 & \textbf{0.56} & \textbf{0.44} & \textbf{0.68} \\
    \cmidrule(r){2-7}
    \textit{P}-GCN                 & 0.59 & \textbf{0.56} & 0.63 & \textbf{0.54} & \textbf{0.47} & 0.62 \\
    \textit{\textbf{I}}-GCN        & \textbf{0.62} & 0.54 & \textbf{0.70} & \textbf{0.54} & 0.33 & \textbf{0.76} \\
    \bottomrule
    \end{tabular}
    \caption{Test-set F$_1$ scores (macro average \textit{Avg}, \textit{D}epressed, \textit{C}ontrol) on DAIC–WOZ and E–DAIC, for participant-only (\textit{P}) and interviewer-only (\textit{I}) variants of Longformer and GCN. For each model, the test result corresponds to the same configuration that achieved the best dev performance in Table~\ref{tab:main_results_three_datasets_dev}.}
    \label{tab:test_results_daic_edaic_nosource}
\end{table}

In Table~\ref{tab:test_results_daic_edaic_nosource} we also report results on the official test sets of DAIC-WOZ and E-DAIC to assess whether the interviewer bias carries over to unseen data. On E-DAIC, both GCN and Longformer models show that interviewer-only variants outperform or match participant-only variants, confirming that the bias persists beyond the dev split. On DAIC-WOZ, the effect is architecture-dependent: keyword-based GCNs still benefit from interviewer turns (I-GCN 0.62 vs P-GCN 0.59), while not in case of the context-based Longformer (P-Longformer 0.68 vs I-Longformer 0.53), suggesting that semantic modeling could partially mitigate prompt-driven shortcuts in some cases.

Overall, these results show that interviewer prompts can inflate performance in semi-structured interviews, independent of architecture. This cross-dataset contrast underscores our central point: gains attributed to ``using interviewer questions as context" ~\cite{chen-naacl-2024,agarwal2024,milintsevich2023towards,shen2022automatic} may simply reflect prompt-driven bias rather than improved modeling of the participant’s language.

\section{Analysis and Discussion}
\label{secc:results}

\begin{figure*}[t]
  \centering
  \begin{minipage}[b]{0.49\textwidth}
    \centering
    \begin{subfigure}[t]{\linewidth}
      \centering
      \includegraphics[width=\linewidth]{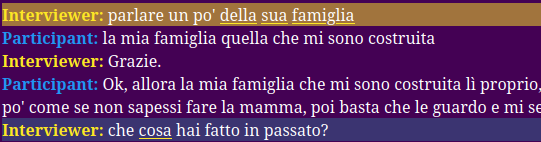}
      \caption{Example interview from ANDROIDS. First utterance translates into `talk about your family'}
      \label{fig:androids-38-pf31-2}
    \end{subfigure}

    \begin{subfigure}[t]{\linewidth}
      \centering
      \includegraphics[width=\linewidth]{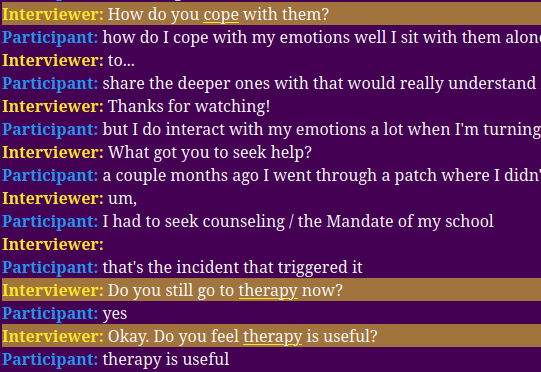}
      \caption{Example of a depressed interview from E-DAIC}
      \label{fig:Edaic_depressed_test_22}
    \end{subfigure}

    \begin{subfigure}[t]{\linewidth}
      \centering
      \includegraphics[width=\linewidth]{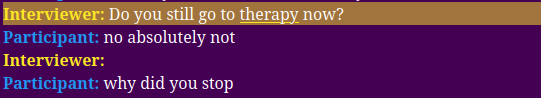}
      \caption{Example of a control interview from E-DAIC}
      \label{fig:Edaic_control_test_26}
    \end{subfigure}
  \end{minipage}\hfill
  \begin{minipage}[b]{0.49\textwidth}
    \centering
    \begin{subfigure}[t]{\linewidth}
      \centering
      \includegraphics[width=\linewidth]{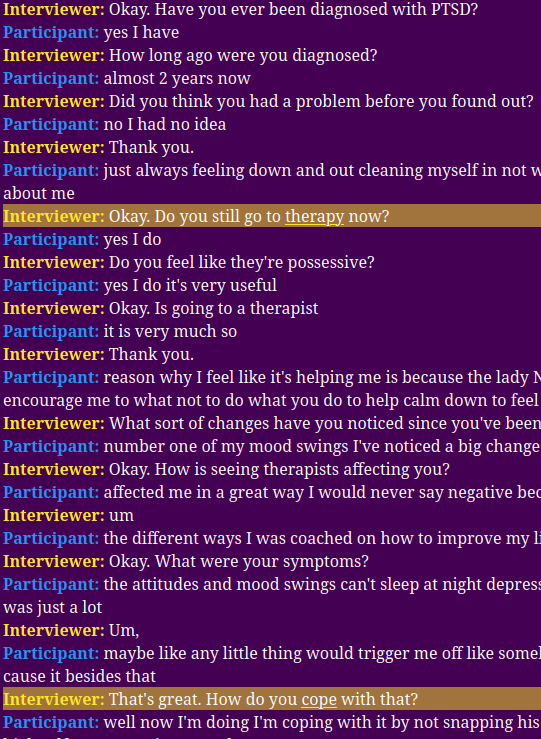}
      \caption{Example of a depressed interview from E-DAIC}
      \label{fig:Edaic_depressed_test_02}
    \end{subfigure}
  \end{minipage}

  \caption{Color-coded interview excerpts in which prompts identified by the \emph{I}-model as bias-carrying are highlighted. Underlined words denote the model’s learned \emph{keywords}, corresponding to the high-contrast narrow bands in Figure~\ref{fig:heatmaps-side-by-side}.} 
  \label{fig:example-interviews}
\end{figure*}

The heatmaps in Figure \ref{fig:heatmaps-side-by-side} visualizes where each model finds decision evidence (words used by the model to identify the depressed group; which we refer as ``keywords") over time and by speaker. For every interview (x-axis), we aggregate the model’s learned “keywords” along the normalized interview timeline (y-axis, 0–100\%) and plot keyword density heatmaps separately for interviewer-only (\textit{I}-GCN) and participant-only (\textit{P}-GCN) models, and for Depressed vs. Control cohorts.

By analyzing the heatmaps across different datasets, we consistently found that patterns diverge sharply by speaker stream. \textit{I}-GCN exhibits narrow, high-contrast bands, indicating that the model relies on specific interviewer turns to make its decision (concentrated keywords). In contrast, \textit{P}-GCN shows low-contrast activity spread across most of the timeline, consistent with drawing evidence from many participant utterances rather than a few fixed positions. The qualitative content of those interviewer bands differs by dataset,

\noindent \textbf{ANDROIDS}: \textit{I}-GCN repeatedly focus on prompts that probe family context, how the last week was spent, and work status (see Fig. \ref{fig:androids-38-pf31-2}). These focused regions recur across interviews even though their absolute timing varies, suggesting the model has learned the type of prompt.

\noindent \textbf{E-DAIC and DAIC-WOZ}: \textit{I}-GCN mainly concentrates on three prompts---“How do you cope with that?”, “Do you still go to therapy?” followed by “Do you feel therapy is useful?”---while largely ignoring other clinically relevant prompts (e.g., PTSD screening, reasons for seeking help, symptoms, and effects of therapy). This again reflects selectivity for a small subset of interviewer turns. Depressed interviews in Fig.~\ref{fig:Edaic_depressed_test_22} and Fig.~\ref{fig:Edaic_depressed_test_02} clearly illustrate this shortcut behavior. Fig.~\ref{fig:Edaic_control_test_26} shows a control subject: after the highlighted bias prompt, the follow-up questioning pattern shifts, providing enough signal for the model to distinguish control from depressed interviews.

\section{Conclusion}
\label{secc:Conclusion}

Our analysis shows that interviewer prompts enable models to distinguish between depressed and control participants across DAIC-WOZ, ANDROIDS, and E-DAIC, with Interviewer models focusing on specific, localized questions whereas Participant models distribute cues broadly across the conversation. These results highlight a key methodological issue: including interviewer turns introduces unintended biases, as models may exploit scripted prompts as shortcuts rather than learning genuine linguistic or behavioral markers of depression. Importantly, this bias is consistent across datasets and model architectures, underscoring its relevance as a general methodological concern rather than an artifact of any particular experimental setup. Future work should carefully account for the interviewer's influence when designing and evaluating conversational mental health assessment systems, for instance by isolating participant-only turns or developing bias-aware evaluation protocols. Without such safeguards, reported gains risk reflecting the consistency of the interview protocol rather than the participant's language.

\section{Acknowledgments}
This work was partially funded by the SNSF through the SPIRIT project \textbf{ORIENTER}: tOwards undeRstanding and modelIng the language of mENTal health disordERs (grant no. IZSTZ0\_223488).

\section{Ethics Statement}

\noindent\textbf{Data privacy, consent, and dataset use.}
Our study relies on publicly distributed clinical-interview corpora with documented consent and privacy safeguards. For DAIC–WOZ and E-DAIC, participants completed informed consent prior to interviews; consent materials included an option permitting data sharing for research. The released transcriptions underwent systematic de-identification (e.g., removal of names, specific dates, addresses) and identifying utterances are withheld; only appropriately anonymized audio/video features and transcripts are distributed under the institutional ethical guidelines, with broader raw data shared case-by-case. For ANDROIDS, data were collected in mental-health centers under institutional and national ethical regulations, with all participation voluntary and every participant signing an informed-consent letter; psychiatrists provided diagnostic labels (DSM-5 framework), and the release includes turn segmentation and person-independent protocols while protecting identity. In our work, we use text only (including ASR where ground truth is unavailable), do not attempt re-identification, and adhere to all dataset usage terms. \\

\noindent\textbf{Role of AI in Healthcare.}
Our experiments are intended to underscore the value of interpretable, AI-assisted methods as decision \emph{support}, not as replacements for clinicians. Diagnostic authority must remain with qualified professionals; delegating clinical decisions to an algorithm introduces unacceptable risk in high-stakes healthcare settings. By exposing shortcut learning and prompt-induced biases in interview data, our work contributes to the development of bias-aware models and evaluation practices for clinical interview analysis.

\section{Limitations}

\noindent\textbf{Transcript Quality and Ground Truth}
ANDROIDS provides no manual transcripts and E-DAIC releases only participant transcripts; complete transcripts for both of these datasets were generated with ASR. Consequently, comparisons between \textit{P} and \textit{I} models can be affected by ASR errors and speaker segmentation, and any advantage/disadvantage may partly reflect transcription noise rather than underlying language alone. Ground-truth, two-speaker transcripts for E-DAIC and ANDROIDS would enable a more precise estimate of prompt-induced bias. \\

\noindent\textbf{Modality Restriction}
Our analysis is text-only across ANDROIDS, DAIC-WOZ, and E-DAIC. While this isolates the linguistic contribution, it does not capture the acoustic and visual features, where interviewer structure and participant signals may manifest differently. As future work, we expect to extend the study to multimodal aspects, running the same \textit{P} vs. \textit{I} ablations to verify whether the prompt-induced bias persists, weakens, or strengthens when non-text modalities are included.

\section{Bibliographical References}\label{sec:reference}

\bibliographystyle{lrec2026-natbib}
\bibliography{main}

\section{Language Resource References}
\label{lr:ref}
\bibliographystylelanguageresource{lrec2026-natbib}
\bibliographylanguageresource{languageresources}

\end{document}